\documentclass[11pt]{article}

\usepackage{acl2023}

\usepackage{times}
\usepackage{latexsym}

\usepackage[T1]{fontenc}

\usepackage[utf8]{inputenc}

\usepackage{microtype}

\usepackage{inconsolata}

\usepackage{linguex}
\usepackage{bussproofs}
\usepackage{latexsym}
\usepackage{stackengine}
\usepackage{tikz}
\usetikzlibrary{shapes,arrows,positioning}
\usepackage{listings}
\usepackage{proof,lscape}
\newcommand{\rulelabelsize}{\scriptsize}
\newcommand{\FA}{\mbox{\rulelabelsize $>$}}

\newcommand{\QC}{\mbox{\rulelabelsize $\mathbf{QC}$}}
\newcommand{\TR}{\mbox{\rulelabelsize $\mathbf{TR}$}}
\newcommand{\bs}{\backslash}
\lstdefinelanguage{yaml}{
  keywords={true,false,null,y,n},
  keywordstyle=\color{darkgreen},
  basicstyle=\ttfamily\small,
  breaklines=true,
  breakatwhitespace=true,
  tabsize=2,
  sensitive=false,
  morecomment=[l]{\#},
  morestring=[b]",
  morestring=[b]'
}
\usepackage{placeins}
\newcommand{\sa}[2]{\stackanchor{$#1$}{$#2$}}

\newcommand{\qsem}{\texttt{QSEM}}
\newcommand{\system}{\texttt{ccg2hol}}

\title{Computational Semantics and Evaluation Benchmark for Interrogative Sentences via Combinatory Categorial Grammar}
\author{Hayate Funakura \\
  Kyoto University\\
  Kyoto, Japan\\
  \texttt{funakura.hayate.28p}\\
  \texttt{@st.kyoto-u.ac.jp} \\\And
  Koji Mineshima \\
  Keio University\\
  Tokyo, Japan\\
  \texttt{minesima@abelard.flet.keio.ac.jp} \\}
\begin{document}
\maketitle
\begin{abstract}
We present a compositional semantics for various types of polar questions and \textit{wh}-questions within the framework of Combinatory Categorial Grammar (CCG).
To assess the explanatory power of our proposed analysis, we introduce a question-answering dataset \qsem\ specifically designed to evaluate the semantics of interrogative sentences. We implement our analysis using existing CCG parsers and conduct evaluations using the dataset. Through the evaluation, we have obtained annotated data with CCG trees and semantic representations for about half of the samples included in \qsem. Furthermore, we discuss the discrepancy between the theoretical capacity of CCG and the capabilities of existing CCG parsers.
\end{abstract}

\section{Introduction}

Interrogative sentences, encompassing various question types, hold a crucial position in the study of syntax and semantics within the field of theoretical linguistics~\cite{dayal2016questions}.
Of particular significance are \textit{wh}-questions, which serve as a benchmark for testing linguistic theories that explore the interface between syntax and semantics, including Categorial Grammar~\cite{steedman1996surface}.
For example, the extraction phenomena involved in \textit{wh}-questions, one of the representative examples in the mismatch between syntax and semantics, provide valuable insights for understanding this interface~\cite{kubota2020type}.
However, despite their importance, the exploration of interrogative sentences within the framework of Categorial Grammar remains relatively underdeveloped
with few exceptions~\cite{vermaat2006logic, xiang2021binding}.

Furthermore, while computational linguistics has witnessed growing research on question sentences in terms of semantic parsing~\cite{kwiatkowski2011lexical,reddy2014},
there exists a notable disparity between the semantic parsing literature and theoretical investigations into the syntax-semantics interface.
The latter research focuses on formal semantics and its detailed examination of various semantic phenomena. This disparity presents an opportunity for bridging the gap and fostering a more integrated approach to the study of interrogative sentences.

Motivated by these gaps in the current literature, this paper aims to present a compositional analysis of different types of interrogatives, including polar and \textit{wh}-questions, within the framework of Combinatory Categorial Grammar (CCG)~\cite{steedman2000syntactic}. This analysis defines a procedure to assign logic-based semantic representations to both questions and their answers, based on their respective CCG trees. These representations can be combined with automated theorem provers to perform logical inferences for question-answering.\footnote{As will be mentioned later, we reduce question-answering to recognizing textual entailment. Therefore, a theorem prover can be used as a question-answering engine.}

To facilitate practical implementation and empirical testing, a computational system \system\ will be introduced in this paper.
This system leverages existing CCG parsers and can be employed for question-answering tasks by integrating it with a theorem prover.

In order to evaluate the syntactic and semantic analyses of interrogative sentences, we design and introduce a dataset of Question-Answer pairs, which we call \texttt{QSEM}.\footnote{\qsem\ is available at \url{https://github.com/hfunakura/qsem}.}
The construction of this dataset follows the methodological approach established by FraCaS ~\cite{cooper1996using}, which serves as a reliable starting point for natural language inferences that carefully separates the semantic and pragmatic factors involved in determining entailment relations. The \qsem\ dataset comprises two primary categories of problems: complex and diverse issues frequently discussed in formal semantics, such as generalized quantifiers and scope ambiguity, and problems that are closer to real-world language use commonly observed in question-answering contexts. The former was created based on the FraCaS problems, while the latter was developed using SQuAD v2.0 \cite{rajpurkar2018know} training data as a basis.
The dataset will provide a valuable resource for detailed examination and analysis of the semantic entailment output by the implemented system.

By undertaking this investigation, we aim to not only contribute to the understanding of interrogative sentences within the context of Categorial Grammar but also shed light on the challenges and limitations of the existing CCG parser based on CCGBank~\cite{hockenmaier-steedman-2007-ccgbank}. Through a thorough examination of interrogative sentences, this study tries to enhance our comprehension of the syntax, semantics, and computational aspects involved, thereby offering valuable insights for future research and applications in the field of computational linguistics and related fields.

What we prioritize most is the establishment of the system \system , which makes theoretical linguistics computationally implementable.
In addition to that, our contributions lie in the following four aspects.

\begin{enumerate}
    \setlength{\itemsep}{0.5pt}
    \setlength{\parskip}{0.5pt}
    \setlength{\parsep}{0.5pt}
    \item We present a compositional analysis that maps various types of interrogative sentences to logical semantic representations within the framework of CCG.
    \item We introduce a FraCaS-inspired benchmark \qsem\ for evaluating the syntax-semantics interface for various types of interrogative sentences.
    \item We report a semantic annotation project
    which assigns each sentence in \qsem\ with a gold CCG tree and a logical semantic representation using our system \system.
    \item We perform a qualitative analysis of the output from standard CCG parsers.
\end{enumerate}

The paper is structured as follows.
In Section 2 we introduce some background in formal semantics and 
computational semantics of interrogative sentences.
In Section 3, we present our analysis of 
the syntax and semantics of interrogative sentences in CCG.
In Section 4, we provide an overview of the \qsem\ dataset and its characteristics.
In Section 5, we introduce our semantic composition and logical inference system \system, which is based on existing CCG parsers and automated theorem provers. We also describe the evaluation and annotation using this system, as well as the qualitative analysis of the CCG parsers.

\section{Related work}
The compositional semantics of interrogative sentences began with \citet{hamblin1973questions} and \citet{karttunen1977syntax}, and has been developed by subsequent researches \cite[etc.]{groenendijk1984studies, krifka2001structured, inquisitiveBook}. Research in this area covers a wide range of topics, including question-answer relationships, presuppositions, and scope problems. Also, various phenomena related to the embedding of interrogative clauses are being actively addressed. While various proposals have been made for specific phenomena and constructions, it is not clear how to test the applicability of each analysis to a wide range of interrogative sentences. It is an important question whether the analyses proposed within a given paper are valid only in a very limited number of cases or whether they have a high degree of generality.

This situation is not limited to questions, but to formal semantics in general. FraCaS~\cite{cooper1996using} is an early benchmark proposed as a basis for systematically evaluating proposals in formal semantics and there have been several subsequent test sets proposed for the evaluation
of formal semantics since then, including MultiFraCaS project\footnote{\url{https://gu-clasp.github.io/multifracas/}} and JSeM \cite{kawazoe2017inference}.\footnote{\url{https://github.com/DaisukeBekki/JSeM}} \citet{watanabe2019question} provide a dataset for evaluating the semantics of questions, including examples of \textit{wh}-questions, polar questions, and alternative questions. However, the dataset has limitations in variation, as it does not include \textit{wh}-words other than \textit{who}, and there are no instances where the object is a \textit{wh}-word. To the best of our knowledge, there is no inference test suite that covers a broader range of linguistic constructions and phenomena related to questions than \qsem.

\section{Syntax and semantics}
We give our analysis to the following types of questions:

\begin{itemize}
    \setlength{\itemsep}{0.5pt}
    \setlength{\parskip}{0.5pt}
    \setlength{\parsep}{0.5pt}
    \item Polar questions
    \item Argument \textit{wh}-questions (\textit{who}, \textit{what}, \textit{which})
    \item Adjunct \textit{wh}-questions (\textit{when}, \textit{where})
\end{itemize}

We seek here to account for the question-answer relationship. Other semantic phenomena associated with questions include presuppositions, ambiguity in question-embedded sentences, and the anaphoric nature of polarity particles. We limit our account here to the following question-response pairs, where the goal is to describe that the response is the answer to the question.

\ex. Polar questions
    \a. Did John meet Mary?
    \b. John met Mary.

\ex. Argument \textit{wh}-questions
    \a. Who smokes?
    \b. John smokes.

\ex. \textit{When}-questions
    \a. When did John meet Mary?
    \b. John met Mary yesterday.

\ex. \textit{Where}-questions
    \a. Where did John meet Mary?
    \b. John met Mary at the station.

We define the relationship between questions and answers in terms of entailment and contradiction relations. In other words, our theory predicts that a response is an answer to a question when the semantic representation of the response entails or contradicts the semantic representation of the question.

The language for semantic representation is a higher-order logic 
language~\cite{mineshima-etal-2015-higher}, combined with event, time, and location variables. Intuitionistic logic is assumed as the logical system, and Coq is used as the inference engine accordingly. Section \ref{subsec-semantic-representations} provides examples of the semantic representations assigned to each type of interrogative. Section \ref{subsec-compositional} discusses the derivation of semantic representations by CCG.

\subsection{Semantic representations for questions}\label{subsec-semantic-representations}
The following are examples of the semantic representations we assign to each type of interrogative sentence.

\ex. Polar questions
    \a. Did John meet Mary?
    \b. $?(\exists x.\exists y.\exists e.[\mathrm{John}(x)\land \mathrm{Mary}(y)\land \mathrm{Meet}(e)\land\mathrm{Subj}(e, x)\land\mathrm{Obj}(e, y)])$

\ex. Argument \textit{wh}-questions
    \a. Who smokes?
    \b. $\mathsf{Q}(\lambda x.\exists e.[\mathrm{Smoke}(e)\land\mathrm{Subj}(e, x)])$

\ex. \textit{When}-questions
    \a. When did John meet Mary?
    \b. $\mathsf{Q}(\lambda t.\exists x.\exists y.\exists e.[\mathrm{John}(x)\land \mathrm{Mary}(y)$\\$\land \ \mathrm{Meet}(e)\land\mathrm{Subj}(e, x)\land\mathrm{Obj}(e, y)$\\ $\land \ \mathrm{TimeOf}(e, t)])$

\ex. \textit{Where}-questions
    \a. Where did John meet Mary?
    \b. $\mathsf{Q}(\lambda l.\exists x.\exists y.\exists e.[\mathrm{John}(x)\land \mathrm{Mary}(y)$\\$\land \ \mathrm{Meet}(e)\land\mathrm{Subj}(e, x)$\\$\land \ \mathrm{Obj}(e, y)\land \mathrm{LocOf}(e, l)])$

What exactly the operators $?$ and $\mathsf{Q}$ should be is a purely semantic question. 

No matter how they are defined, there is no effect on semantic composition. Since our goal is to establish a semantic composition workflow consistent with the CCG parsers, we define $?$ and $\mathsf{Q}$ in a very simple form.

\ex.
\a. \label{def:?} $?(P)\equiv P\lor\lnot P$ 
\b.[] \hfill (where $P$ is a formula of type $t$)
\c.\label{def:Q} $\mathsf{Q}(f)\equiv \exists x.f(x)$ 
\d.[] \hfill (where $f$ is a first-order function)

The above representations are partially based on those of inquisitive semantics~\cite{inquisitiveBook};
for polar questions they are the same as in inquisitive semantics,
while for wh-questions, we discard the ambiguity about exhaustivity that is considered in \citet{inquisitiveBook},
thus simplifying the treatment of inquisitive semantics.

There are various alternative options for defining $?$ and $\mathsf{Q}$.
We mention two of them.
First, a Karttunen-style analysis can be achieved by defining these operators as follows:

\ex. $?(P)\equiv \lambda p.[p(w_{a})\land p=\lambda w.p(w)]$

\ex. $\mathsf{Q}(f)\equiv$ \\ $\lambda p.\exists x.[p(w_{a})\land p=\lambda w.f(x)(w)]$

Here, $w_{a}$ denotes a designated (actual) world.

Second, it is also possible to define the $?$ and $\mathcal{Q}$ operators 
in terms of modal logic, which enables to express three readings with respect to exhaustivity by providing three $\mathcal{Q}$ operators~\cite{nelken2004logic,nelken2006modal}.

\ex. Semantic representation of questions using modality
\a. $\mathsf{Q}_{ms}(f) = \exists x.[\Box f(x)]$ \b.[] \hfill (mention-some reading)
\c. $\mathsf{Q}_{we}(f) = \forall x.[f(x)\to\Box f(x)]$
\d.[] \hfill (weakly exhaustive reading)
\e. $\mathsf{Q}_{se}(f) = \forall x.[(f(x)\to\Box f(x))\land (\lnot f(x)\to\Box \lnot f(x))]$
\f.[] \hfill (strongly exhaustive reading)

As discussed above, depending on how $?$ or $\mathsf{Q}$ are defined, this analysis can embody various perspectives. We do not intend to commit to a specific position. Therefore, we adopt \ref{def:?} and \ref{def:Q} for simplicity. Note that we have chosen intuitionistic logic as the underlying logic, mainly because of its compatibility with theorem provers.

\subsection{Compositional semantics}\label{subsec-compositional}
In this subsection, we present the lexical items defined for the words that play a central role in our analysis: \textit{be}, \textit{do}, and \textit{wh}-words. We also demonstrate semantic composition using them.

\subsubsection{Lexical entries}
\begin{table*}[h]
\centering
\scalebox{0.7}{
\begin{tabular}{c|c|c}
Expression & Category & Semantics \\ \hline
$\mathrm{be}_{1}$ & $(S_{q}/NP)/NP$ & $\lambda P_{1}P_{2}K.P_{2}(\lambda y.\top, \lambda x.Q_{1}(\lambda y.\top, \lambda y.\exists e.[\mathrm{Be}(e)\land (\mathrm{Subj}(e)=y)\land K(e)])$\\
$\mathrm{be}_{2}$ & $(S_{q}/(S_{adj}\bs NP))/NP$ & $\lambda P_{1}P_{2}K.P_{2}(\lambda y.\top, \lambda x.P_{1}(\lambda y.\top, \lambda y.\exists e.[\mathrm{Be}(e)\land (\mathrm{Subj}(e)=y)\land K(e)]))$ \\
$\mathrm{be}_{3}$ & $(S_{q}/(S_{pss}\bs NP))/NP$ & $\lambda P_{1}P_{2}K. P_{2}(P_{1}, \lambda e. K(e))$ \\
$\mathrm{do}$ & $(S_{q}/(S_{b}\bs NP))/NP$ & $\lambda P_{1}P_{2}K. P_{2}(P_{1},K)$
\end{tabular}
}
\caption{Lexical entries for \textit{be} and \textit{do}}
\label{tab:lexbedo}
\end{table*}

\begin{table*}[h]
\centering
\scalebox{0.7}{
\begin{tabular}{c|c|c}
Expression & Category & Semantics \\ \hline
$\mathrm{who}$ &
$S_{wq}/(S|NP)$ &
$\lambda PK.\mathsf{Q}(\lambda x.P(\lambda F_{1}F_{2}.F_{2}, \lambda y.\top))$ \\
$\mathrm{what}_{1}$ &
$S_{wq}/(S|NP)$ &
$\lambda PK.\mathsf{Q}(\lambda x.P(\lambda F_{1}F_{2}.F_{2}, \lambda y.\top))$ \\
$\mathrm{what}_{2}$ & $(S_{wq}/(S|NP))/N$ & $\lambda P_{1}P_{2}K.\mathsf{Q}(\lambda x.[P_{1}(x)\land P_{2}(\lambda F_{1}F_{2}.F(x), \lambda y.\top)])$\\
$\mathrm{which}$ & $(S_{wq}/(S|NP))/N$ & $\lambda P_{1}P_{2}K.\mathsf{Q}(\lambda x.[P_{1}(x)\land P_{2}(\lambda F_{1}F_{2}.F(x), \lambda y.\top)])$\\
$\mathrm{when}_{1}$ & $S_{wq}/S_{q}$ & $\lambda SK.\exists t.\mathsf{Q}(S(\lambda e.\mathrm{TimeOf(e, t)}))$\\
$\mathrm{when}_{2}$ & $S_{wq}/(S_{q}/NP)$ & $\lambda PK.\mathsf{Q}(\lambda t.P(\lambda F_{1}F_{2}.(F_{1}\land F_{2}), \lambda e.\exists t.\mathrm{TimeOf}(e, t)))$\\
$\mathrm{where}_{1}$ & $S_{wq}/S_{q}$ & $\lambda SK. \exists l.\mathsf{Q}(S(\lambda e.\mathrm{LocOf}(e, l)))$\\
$\mathrm{where}_{2}$ & $S_{wq}/(S_{q}/NP)$ & $\lambda PK.\mathsf{Q}(P(\lambda F_{1}F_{2}.(F_{1}\land F_{2}), \lambda e.\exists l.\mathrm{LocOf}(e, l)))$
\end{tabular}
}
\caption{Lexical entries for \textit{wh}-expressions. Here, we bundle $S_{dcl}\bs NP$ and $S_{q}/NP$ together and denote them as $S|NP$.}
\label{tab:lexwh}
\end{table*}

\begin{figure*}[h]
    \centering
    \scalebox{0.8}{
    \infer[\QC]{\bar{S}_{wq}:\mathsf{Q}(\lambda x.\exists e.[\mathrm{Smoke}(e)\land\mathrm{Subj}(e, x)])}{
    \infer[\FA]{S_{wq}:\lambda K.\mathsf{Q}(\lambda x.\exists e.[\mathrm{Smoke}(e)\land\mathrm{Subj}(e, x)])}{
        \infer{\stackanchor{$S_{wq}/(S_{dcl}\bs NP)$}{$\lambda PK.\mathsf{Q}(\lambda x.P(\lambda F_{1}F_{2}.F_{2}, \lambda y.\top))$}}{\mbox{Who}} &
        \infer{\stackanchor{$S_{dcl}\bs NP$}{$\lambda PK.P(\lambda x.\top, \lambda x.\exists e.[\mathrm{Smoke}(e)\land\mathrm{Subj}(e, x)\land K(e)])$}}{
            \mbox{smokes}
            }
        }
    }
    }
    \caption{An example of semantic composition for a \textit{wh}-question. To ensure Categorial Type Transparency \cite{steedman2000syntactic}, the category derived by $\mathbf{QC}$ is distinguished from $S_{wq}$ and is denoted as $\bar{S}_{wq}$.}
    \label{fig:wh-semcomp}
\end{figure*}

\begin{figure*}[h]
    \centering
    \scalebox{0.7}{
    \infer[$?I$]{\sa{\bar{S}_{pol}}{?\exists e [\mathrm{Like}(e)\land (\mathrm{Subj}(e)=\mathrm{John})\land (\mathrm{Obj}(e)=\mathrm{Smith})]}}{
    \infer[\QC]{\sa{\bar{S}_{q}}{\exists e [\mathrm{Like}(e)\land (\mathrm{Subj}(e)=\mathrm{John})\land (\mathrm{Obj}(e)=\mathrm{Smith})]}}{
    \infer[\FA]{\sa{S_{q}}{\lambda K. \exists e [\mathrm{Like}(e)\land (\mathrm{Subj}(e)=\mathrm{John})\land (\mathrm{Obj}(e)=\mathrm{Smith}) \land K(e)]}}{
    \infer[\FA]{\sa{S_{q}/(S_{b}\backslash NP)}{\lambda P_{2}K.Q_{2}(\lambda F_{1}F_{2}.F_{1}(\mathrm{John})\land F_{2}(\mathrm{John}), K)}}{
    \infer{\sa{(S_{q}/(S_{b}\backslash NP))/NP}{\lambda P_{1}P_{2}K.Q_{2}(Q_{1}, K)}}{\mbox{Does}} & 
    \infer[\TR]{\sa{NP}{\lambda F_{1}F_{2}.F_{1}(\mathrm{John})\land F_{2}(\mathrm{John})}}{
    \infer{\sa{N}{\mathrm{John}}}{\mbox{John}}}} & 
    \infer[\FA]{\sa{S_{b}\backslash NP}{\lambda P_{2}K.P_{2}(\lambda y.\top, \lambda x. \exists e. [\mathrm{Like}(e)\land (\mathrm{Subj}(e)=x)\land (\mathrm{Obj}(e)=\mathrm{Smith}) \land K(e)])}}{
    \infer{\sa{(S_{b}\backslash NP)/NP}{\cdots }}{\mbox{like}} & 
    \infer[\TR]{\sa{NP}{\lambda F_{1}F_{2}.F_{1}(\mathrm{Smith})\land F_{2}(\mathrm{Smith})}}{
    \infer{\sa{N}{\mathrm{Smith}}}{\mbox{Smith}}}}}
    }
    }
    }
    \caption{An example of semantic composition for a polar question}
    \label{fig:pol-semcomp}
\end{figure*}

The lexical entries for \textit{be} and \textit{do} are shown in Table \ref{tab:lexbedo}.\footnote{We have not attributed the $?$ operator appearing in the semantic representation of polar questions to the lexical meaning of \textit{be} or \textit{do}, but have defined a unary rule of CCG that introduces the $?$ operator. This is more of a practical measure for ease of implementation rather than a theoretical one.} We assume that \textit{be} and \textit{do} appearing in interrogative sentences are distinct in the lexicon from those appearing in declarative sentences. In other words, separate lexical entries are defined for \textit{be} and \textit{do} that appear in declarative sentences (omitted here).

The lexical entries for \textit{wh}-words are shown in Table \ref{tab:lexwh}. The morphemes $\mathrm{when}_{2}$ and $\mathrm{where}_{2}$ are assumed to appear in the construction \texttt{wh-be-NP} (see examples below).

\ex. When is the deadline?

\ex. Where is the office?

\subsubsection{Semantic composition}
In this subsection, we provide examples of semantic composition for both \textit{wh}-questions and polar questions.

Figure \ref{fig:wh-semcomp} shows an example of semantic composition for a \textit{wh}-question. Since we assume that the verb introduces event quantification, we use the Quantifier Closure rule ($\mathbf{QC}$) in addition to the CCG combinatory rules. $\mathbf{QC}$ is a unary rule, which applies the input expression to $\lambda x.\top$.

An example of semantic composition for polar questions is shown in Figure \ref{fig:pol-semcomp}. To derive the semantic representation of polar questions, we define a CCG unary rule, $\mathbf{?I}$, which is a rule that transforms an expression $P$ into $?P$, through which the semantic representation of a polar question is obtained.

\section{Dataset}
\texttt{QSEM} consists of questions and the responses to those questions. The primary goal in creating this dataset was to establish a basis for quantitatively measuring the degree of agreement between the predictions derived from semantic representations and our intuitive judgments. What this dataset asks the system is whether a given answer qualifies as an answer to the question. 

The following are examples of the samples included in the dataset. ``P'' represents the premise and ``Q'' represents the question. Labels have three possible values: \texttt{yes}, \texttt{no}, and \texttt{unknown} (the rules for label assignment are discussed in Section \ref{sec:process}).

\ex. ID: 6
    \a.[P1] Every Italian man wants to be a great tenor.
    \b.[Q] Who wants to be a great tenor?
    \c.[Label: ] yes

\ex. ID: 35
    \a.[P1] No delegate finished the report on time.
    \b.[Q] Which delegate finished the report on time?
    \c.[Label: ] no

\ex. ID: 72
    \a.[P1] Amish separated from the Mennonites in 1693.
    \b.[Q] When did the Anabaptists split?
    \c.[Label: ] unknown

The format of our dataset is based on FraCaS \cite{cooper1996using}, a pioneering semantic evaluation dataset.
Before going into the details of our dataset,
the following subsection provides an overview of FraCaS as a background.

\subsection{Background: FraCaS}
FraCaS is a test suite for evaluating NLP systems and linguistic theories. The first version provided in \citet{cooper1996using} contains one or more assumptions, a polar question, and an answer to that question (\textit{yes}, \textit{no}, \textit{don't know}, etc.). There are 346 problems, divided into sections for each phenomenon. And it is controlled not to include difficulties other than the phenomenon in focus. This makes it easy to estimate the explanatory power of the analysis by phenomenon.

The following is an example of an original problem:

\ex. ID: 3.1 in \citet{cooper1996using}
    \a.[P1] An Italian became the world's greatest tenor.
    \b.[Q] Was there an Italian who became the world's greatest tenor?
    \c.[Label: ] yes

The original form was thus a dataset consisting of QA pairs, but then hypotheses (H) were added by Bill MacCartney and formulated as a set of implication recognizing textual entailment.\footnote{This version is available at \url{https://nlp.stanford.edu/~wcmac/downloads/fracas.xml}} The following are examples of questions from the latest version of FraCaS:

\ex. ID: fracas-001
    \a.[P1] An Italian became the world's greatest tenor.
    \b.[Q] Was there an Italian who became the world's greatest tenor?
    \c.[H] There was an Italian who became the world's greatest tenor. 
    \d.[Label: ] yes

\ex. ID: fracas-085
    \a.[P1] Exactly two lawyers and three accountants signed the contract.
    \b.[Q] 	Did six lawyers sign the contract?
    \c.[H] Six lawyers signed the contract.
    \d.[Label: ] no

\ex. ID: fracas-117
    \a.[P1] Every student used her workstation.
    \b.[P2] Mary is a student.
    \c.[Q] Did Mary use her workstation?
    \d.[H] Mary used her workstation.
    \e.[Label: ] yes

FraCaS includes a wide range of topics, including tense, anaphora, and propositional attitudes. In addition, all samples include polar questions. However, there is no section focusing on the semantic behavior of the questions themselves. Also, to our knowledge, there are few other evaluation datasets for question semantics. This motivates our proposed dataset. In the following subsections, we describe the contents of our dataset and the process of its creation.

\subsection{Dataset organization}
\begin{table}[t]
\centering
\begin{tabular}{|l|c|}
\hline
\textbf{Type of Question} & \textbf{Count} \\
\hline
Polar & 23 \\
Who & 15 \\
What & 20 \\
Which & 22 \\
When & 35 \\
Where & 23 \\
\hline
\end{tabular}
\caption{The number of samples for each type of question}
\label{tab:questionRatio}
\end{table}

Our dataset consists of 138 samples. We place more emphasis on the qualitative aspects, such as the variety of constructions and phenomena, and the accuracy of labels, rather than quantitative aspects.

Table \ref{tab:questionRatio} shows the number of samples for each type of question. Each sample is annotated as to which type of question is it related to. Thus, it is easy to measure the degree to which the system is applicable to which type of questions. A wider range of questions, such as alternative questions, \textit{how}-questions, \textit{why}-questions, etc., will need to be covered in the future. 

The dataset consists of the following problems:

\begin{enumerate}
    \setlength{\itemsep}{0.5pt}
    \setlength{\parskip}{0.5pt}
    \setlength{\parsep}{0.5pt}
    \item\label{prob:GQ} Problems that test understanding of quantificational expressions
    \item\label{prob:multiple} Problems that test syntactic and semantic understanding of multiple \textit{wh}-questions
    \item\label{prob:scope} Problems that test understanding of the interaction of quantifiers and \textit{wh}-word scopes
    \item\label{prob:basic} Problems that test comprehension of basic \textit{wh}-questions
\end{enumerate}
\ref{prob:GQ}-\ref{prob:scope} focuses on whether the system can solve semantically challenging problems. \ref{prob:basic}, on the other hand, focuses on the general applicability of proposed analyses. Each of the above four types of questions was created from different resources. In the following subsection, we will explain the process of creating each question, presenting sample examples.

\subsection{Dataset creation process}\label{sec:process}
In this subsection, we describe the labeling rules and how we collected the QA pairs.

\paragraph{Labeling rules}

Each QA pair in \qsem\ is assigned one of the labels: \texttt{yes}, \texttt{no}, or \texttt{unknown}. The rules for label assignment are as follows.

\begin{itemize}
    \setlength{\itemsep}{0.5pt}
    \setlength{\parskip}{0.5pt}
    \setlength{\parsep}{0.5pt}
    \item Problems copied from FraCaS
    \begin{itemize}
        \item We use the original labels assigned in FraCaS.
    \end{itemize}
    \item Problems created by the authors
    \begin{itemize}
        \item When the premises directly answer the question, \texttt{yes} is assigned.
        \item When the premises negate the presupposition of the question, \texttt{no} is assigned.
        \item When none of the above conditions apply, \texttt{unknown} is assigned.
    \end{itemize}
\end{itemize}

\paragraph{Quantificational expressions}
Pairs of polar questions and responses were extracted from sections 1.1 and 1.2 of FraCaS as samples for quantification. The current version includes only those examples in which either \textit{every}, \textit{all}, \textit{each}, \textit{some}, \textit{a}, or \textit{no} is used. Of the 23 polar questions, 14 were copied from FraCaS. The other 9 were created by the authors based on the FraCaS samples.

\paragraph{Multiple \textit{wh}-questions} The interrogative sentences in which both a fronted \textit{wh}-phrase and a \textit{wh}-phrase in-situ occur together, i.e., multiple \textit{wh}-questions, is also included in this dataset. This construction is ambiguous between single-pair reading and pair-list readings. There are vigorous debates to explain this ambiguity.

The QA pairs on multiple \textit{wh}-questions are taken from \citet{dayal2016questions}. The collected samples are shown below.

\ex. ID: 40
    \a.[P1] Bill met Carl.
    \b.[P2] Bill is a student.
    \c.[P3] Carl is a professor.
    \d.[Q] Which student met which professor?
    \e.[Label: ] yes

\ex. ID: 41
    \a.[P1] Bill met Carl and Alice met Dan.
    \b.[P2-3\quad] Bill is a student., Alice is a student.
    \c.[P4-5\quad] Carl is a professor., Dan is a professor.
    \d.[Q] Which student met which professor?
    \e.[Label: ] yes

P1 and Q are copied from the source literature, while the premises after P2 are added by the authors to provide a context.

\paragraph{Scope ambiguity} \textit{wh}-interrogatives in which quantificational expressions occur are also a central subject of study in this area. 

\ex. Who does everyone like?
    \a.\label{scope:reading:forall-wh} Tell me about one person who is liked by all. \hfill (wh>$\forall$)
    \b.\label{scope:reading:wh-forall} For each person, tell me who that person likes. \hfill ($\forall$>wh)

On the other hand, such ambiguity does not arise when the quantificational expression appears in the object position.

\ex. Who likes everyone? \hfill (wh>$\forall$)

Samples for scope ambiguity were taken from \citet{chierchia1993questions} and \citet{krifka2003quantifiers}. Examples are shown below.

\ex. ID: 48
    \a.[P1] Bill likes Smith and Sue likes Jones.
    \d.[Q] Who does everyone like?
    \e.[Label: ] yes

\ex. ID: 49
    \a.[P1] Everyone likes Smith.
    \d.[Q] Who does everyone like?
    \e.[Label: ] yes

\ex. ID: 44
    \a.[P1] Bill likes Smith and Alice likes Jones.
    \b.[P2-3\quad] Bill is a student., Alice is a student.
    \c.[P4-5\quad] Smith is a professor., Jones is a professor.
    \d.[Q] Which student likes every professor?
    \e.[Label: ] no

\paragraph{Basic \textit{wh}-questions}
\begin{table}[t]
\centering
\begin{tabular}{|c|c|}
\hline
\textbf{Type of Question} & \textbf{Count} \\
\hline
Who & 10 \\
What & 10 \\
Which & 8 \\
When & 35 \\
Where & 23 \\
\hline
\end{tabular}
\caption{The number of samples obtained from SQuAD by question type}
\label{tab:questionRatioSquad}
\end{table}
To test for a greater variety of constructions, samples were created based on SQuAD training data. SQuAD is a question-answering dataset that is often used as a benchmark for natural language processing systems. The dataset contains approximately 90K QA pairs, and the system must provide an answer to a question based on the content of a given paragraph. In addition to the above samples, there are about 40K questions for which no answer can be found from the given paragraphs.

We performed random sampling by question type from the questions in this dataset. From them, we excluded samples in which the following phenomena and constructions were critically involved.

\begin{itemize}
    \setlength{\itemsep}{0.5pt}
    \setlength{\parskip}{0.5pt}
    \setlength{\parsep}{0.5pt}
    \item idiom
    \item coordination
    \item anaphora
    \item tense
    \item degree
\end{itemize}
We also excluded questions on sensitive topics. We performed the above work with a random sampling size of 50 for the \textit{when} and \textit{where} questions, and 30 for the \textit{who}, \textit{what}, and \textit{which} questions. The answers included in SQuAD are so-called non-sentential answers. Based on these, we created answers for \texttt{QSEM}. Table \ref{tab:questionRatioSquad} shows the final sample size obtained for each type of interrogative.

\paragraph{Limitations}
Here, we will discuss the main limitations of \qsem.

\qsem\ includes examples related to negative interrogative sentences (as shown below), but it does not capture the fact that such questions often receive rhetorical interpretations.

\ex. ID: 115
    \a.[P1] Constantius did not consent to a new trial.
    \d.[Q] Who did not consent to a new trial?
    \e.[Label: ] yes

Additionally, while focus plays a crucial role in the relationship between a question and its answer, the current \qsem\ abstracts away from information related to focus.

\section{Implementation and semantic annotation}
This section provides an overview of the annotation and the results of the evaluation.
\subsection{Pipeline}
In this subsection, we describe our system \system\ for semantic composition and logical inference. The pipeline for this system is shown in Figure \ref{fig:pip-of-ccg2hol}. Among the components in the diagram, what we implemented is the semantic composition system that takes CCG derivation trees as input and outputs formulas (HOLs), and the interface between each component.

The system first takes one or more sentences as input and performs syntactic parsing using existing CCG parsers. C\&{C} parser \cite{clark2007wide}, EasyCCG \cite{lewis2014ccg}, and depccg \cite{yoshikawa-etal-2017-ccg} are used as CCG parsers. Based on the results of the parsing, a semantic tag is assigned to each word. Then, the semantic composition is performed using the CCG tree and semantic tags~\cite{abzianidze2017towards}.\footnote{While we have followed the idea of \citep{abzianidze2017towards} to use semantic tags as a key to determine lexical meanings, the specific design of the tag set was carried out by ourselves.} As a semantic representation, we propose an abstract expression that is independent of specific semantic analysis. We call this expression HOL (higher-order logic). HOL has information on syntactic dependencies and semantic tags (e.g., Figure \ref{fig:example-hol}). At present, we are mechanically assigning semantic tags from CCG categories, POS tags, NER tags, and lemmas. Semantic tags were proposed as a superior resource for judging lexical meaning than POS tags or NER tags, so assigning semantic tags based on these pieces of information is not ideal. Therefore, it is desirable for semantic tags to be determined by an independent assigner. In this study, as a provisional measure, we manually supplemented areas where POS tags or NER tags were insufficient (see Section \ref{ssec:evaluation} below).

HOL can be converted into specific expressions such as FOL, DRT, etc. Therefore, by utilizing HOL as the primary semantic representation, our system can be used independently of any specific theoretical framework or analytical strategy. An example of HOL is shown in Figure \ref{fig:example-hol}. HOLs are then converted into FOL expressions for inference. The FOL expressions are passed to the theorem prover Coq \cite{bertot2004interactive} to perform inference and predict entailment and contradiction relations.

\begin{figure*}[t]
    \centering
    $((\mathrm{which}_{\mathsf{WDT}}\ \mathrm{delegate}_{\mathsf{NN}})\ (\mathrm{finish}_{\mathsf{TV}}\ (\mathrm{the}_{\mathsf{DT}}\ \mathrm{report}_{\mathsf{NN}})))$
    \caption{HOL corresponding to \textit{Which delegate finished the report?}}
    \label{fig:example-hol}
\end{figure*}

\subsection{Evaluation and Annotation}
\label{ssec:evaluation}

Using the pipeline described above, we evaluated the degree to which our analysis could address the \qsem\ problem. And we have accumulated samples that contain no errors in HOLs or inference results as gold data. The main source of errors is in deriving HOLs. These errors mainly fall into two categories: mistakes in CCG trees and inaccuracies in semantic tag assignment. If an error could be resolved by simply adjusting the semantic tags, we manually made the corrections.

\begin{figure*}
    \centering
    \begin{tikzpicture}[
      node distance = 4cm,
      auto,
      every node/.style={align=center},
      auxiliary/.style={rectangle,draw=black,fill=blue!20,align=center},
      main/.style={rectangle,draw=black,fill=red!20,align=center}
    ]
    
      \node[main] (sentence) {Sentences};
      \node[main,right = 2cm of sentence] (ccg) {CCG derivation tree};
      \node[auxiliary,below = 0.5cm of ccg] (semtag) {\texttt{semtag.yaml}};
      \node[main,right = 1cm of ccg] (hol) {HOL};
      \node[auxiliary,below = 0.5cm of hol] (univsem) {\texttt{univsem.yaml}};
      \node[main,right = 1cm of hol] (fol) {FOL};
      \node[right = 2cm of fol] (result) {entailment\\ contradiction\\ unknown};

      \draw[->] (sentence) -- node[midway,above] {\scriptsize CCG parser} (ccg);
      \draw[->] (ccg) --  (hol);
      \draw[->] (hol) -- (fol);
      \draw[->] (fol) -- node[midway,above] {\scriptsize Prover} (result);
    
      \draw[->,dashed] (semtag) -- (hol);
      \draw[->,dashed] (univsem) -- (fol);

\end{tikzpicture}
    \caption{Pipeline of \texttt{ccg2hol}}
    \label{fig:pip-of-ccg2hol}
\end{figure*}

In the following, we will discuss the manually assigned semantic tags and provide a qualitative error analysis of the CCG parsers. Lastly, we will report on the extent of completed annotations within the \qsem\ data.

\paragraph{Semantic tags}
To represent the polysemy of prepositions, we manually corrected the output of the system. Using only the information utilized in the above-mentioned pipeline for semantic tagging, we could not differentiate between prepositions used for time and those used for location, leading to the same semantic tag being assigned in all cases. This resulted in difficulties when dealing with examples of \textit{when} and \textit{where} questions. Therefore, we manually assigned different semantic tags to time prepositions appearing in expressions like \textit{on December 12} and location prepositions appearing in expressions like \textit{in Oxford}.

\paragraph{Main errors in CCG trees}
Through the observation of our analysis results, the following tendencies in existing CCG parsers were suggested:

\begin{itemize}
    \setlength{\itemsep}{0.5pt}
    \setlength{\parskip}{0.5pt}
    \setlength{\parsep}{0.5pt}
    \item Multiple \textit{wh}-questions are particularly difficult for parsers.
    \item There is a tendency for the analysis of past participles to be inconsistent between declarative and interrogative sentences.
\end{itemize}

The current version of \qsem\ includes one instance of a multiple-\textit{wh} question. 

\ex. 
ID: 40, 41 \\
Which student met which professor?

All parsers we employed failed in analyzing this instance. C\&{C} parser and depccg identified \textit{which professor} as an embedded interrogative clause. EasyCCG analyzed the two \textit{which} as if they were adjectives, and recognized the entire sentence as a declarative sentence.

In addition, discrepancies were observed between declarative and interrogative sentences regarding past participles appearing as complements to \textit{be}, such as \textit{located} in the following example.

\ex. ID: 129
\a.[Q] Where is Symphony Hall located? 
\b.[A] Symphony Hall is located on the west of Back Bay. 

There are 16 instances in \qsem\ that involve the use of \textsf{be-V$_{\sf pp}$}. For 12 of these, both C\&{C} and depccg recognized the past participle form appearing in interrogatives as an adjective, while recognizing the past participle form appearing in declaratives as the passive voice of a verb. Even in cases where the analysis of the past participle was consistent, there were errors in other parts of the tree. EasyCCG was consistent, recognizing both types of past participle as likely being in the passive voice. However, there was not a single instance where the entire tree was correctly parsed.

\paragraph{Results}
We performed \system\ analysis for each sample in \qsem, and considered those that correctly produced CCG trees, semantic representations, and inference results as gold data for annotation. Currently, annotations have been completed for approximately 49.3\% (68 out of 138) of the entire \qsem. Many of the analyses yet to be annotated include results with parsing errors from the CCG parsers as mentioned above, and results with inaccurate semantic tags assigned.

\section{Conclusion}
In this paper, we proposed an extensive analysis of the interrogative sentences and proposed a benchmark \texttt{QSEM} to evaluate it. In addition, we introduced a system, \system, to implement the proposed analysis. This system was used to annotate a portion of the examples in \qsem\ with CCG trees and HOL.

\texttt{QSEM} aims to formulate interesting problems in question semantics as question-answering and will be further augmented in the future. \texttt{ccg2hol} is a semantic composition and inference system. The HOL obtained as a result of semantic composition is an abstract structure independent of any specific analysis and, together with the annotated data, can be used for testing various syntactic and/or semantic frameworks.

Our ultimate goal is for \system\ to be a language-universal and analysis-independent computational framework. 
To make \system\ a universal inference and evaluation framework, broader annotations and extensions to other languages are necessary. For wider annotation and an improved inference system, the immediate future challenges to tackle are the elimination of parsing errors by the CCG parsers and the refinement of semantic tag design. 
For \system\ to handle other languages,, it is necessary to connect the semantic composition system to parsers for languages other than English.

\section*{Acknowledgment}
This work is partially supported by JST, CREST grant number JPMJCR2114.

\bibliographystyle{acl_natbib}
\bibliography{custom}
\end{document}